\documentclass[sigconf, authorsversion]{acmart}
\usepackage{times}
\usepackage{epsfig}
\usepackage{graphicx}
\usepackage{amsmath}
\usepackage{csquotes}
\usepackage{caption}
\usepackage{caption}
\usepackage{subfig}
\usepackage{textcomp}
\usepackage{booktabs}
\usepackage[hang,flushmargin]{footmisc}
\setcopyright{none}
\def\BibTeX{{\rm B\kern-.05em{\sc i\kern-.025em b}\kern-.08emT\kern-.1667em\lower.7ex\hbox{E}\kern-.125emX}}

\setcopyright{none}
\renewcommand\footnotetextcopyrightpermission[1]{}
\begin{document}

\title{Saliency-Aware Class-Agnostic Food Image Segmentation}

\author{Sri Kalyan Yarlagadda}
\email{yarlagad@purdue.edu}
\affiliation{
    \institution{Purdue University}
    \city{West Lafayette}
    \country{USA}
    }

\author{Daniel Mas Montserrat}
\email{dmasmont@purdue.edu}
\affiliation{
    \institution{Purdue University}
    \city{West Lafayette}
    \country{USA}
    }

\author{David G\"{u}era}
\email{dgueraco@purdue.edu}
\affiliation{
    \institution{Purdue University}
    \city{West Lafayette}
    \country{USA}
    }
    
\author{Carol J. Boushey}
\email{cjboushey@cc.hawaii.edu}
\affiliation{
    \institution{University of Hawaii Cancer Center}
    \city{Honolulu}
    \country{USA}}

\author{Deborah A. Kerr}
\email{d.kerr@curtin.edu.au}
\affiliation{
    \institution{Curtin University}
    \city{Perth}
    \country{Australia}
    }

\author{Fengqing Zhu}
\email{zhu0@purdue.edu}
\affiliation{
    \institution{Purdue University}
    \city{West Lafayette}
    \country{USA}
    }
    
\renewcommand{\shortauthors}{Yarlagadda, et al.}

\begin{abstract}
Advances in image-based dietary assessment methods have allowed nutrition professionals and researchers to improve the accuracy of dietary assessment, where images of food consumed are captured using smartphones or wearable devices. These images are then analyzed using computer vision methods to estimate energy and nutrition content of the foods. Food image segmentation, which determines the regions in an image where foods are located, plays an important role in this process. Current methods are data dependent, thus cannot generalize well for different food types. To address this problem, we propose a class-agnostic food image segmentation method. Our method uses a pair of eating scene images, one before start eating and one after eating is completed. Using information from both the before and after eating images, we can segment food images by finding the salient missing objects without any prior information about the food class. We model a paradigm of top down saliency which guides the attention of the human visual system (HVS) based on a task to find the salient missing objects in a pair of images. Our method is validated on food images collected from a dietary study which showed promising results.
\end{abstract}

\keywords{food segmentation, image based dietary assessment}

\maketitle

\section{Introduction} 
\label{intro}
It is well-known that dietary habits have profound impacts on the quality of one's health and well-being~\cite{Nordstrom2013, Mesas2012}. While a nutritionally sound diet is essential to good health~\cite{organization_2009}, it has been established through various studies that poor dietary habits can lead to many diseases and health complications. For example, studies from the World Health Organization (WHO)~\cite{organization_2009} have shown that poor diet is a key modifiable risk factor for the development of various non-communicable diseases such as heart disease, diabetes and cancers, which are the leading causes of death globally~\cite{organization_2009}. In addition, studies have shown that poor dietary habits such as frequent consumption of fast food~\cite{fast-food}, diets containing large portion size of energy-dense foods~\cite{Piernas2011}, absence of home food~\cite{homefood_absence} and skipping breakfast~\cite{breakfast} all contribute to the increasing risk of overweight and obesity. 
Because of the many popular diseases affecting humans are related to dietary habits, there is a need to study the relationship between our dietary habits and their effect on our health.

Understanding the complex relationship between dietary habits and human health is extremely important as it can help us mount intervention programs to prevent these diet related diseases~\cite{daugherty-jmir2012}. 
To better understand the relationship between our dietary habits and human health, nutrition practitioners and researchers often conduct dietary studies in which participants are asked to subjectively assess their dietary intake. In these studies, participants are asked to report foods and drinks they consumed on a daily basis over a period of time. 
Traditionally, self-reporting methods such as 24-hr recall, dietary records and food frequency questionnaire (FFQ) are popular for conducting dietary assessment studies~\cite{Shim2014}. However, these methods have several drawbacks. For example, both the 24-hr recall and FFQ rely on the participants' ability to recall foods they have consumed in the past. In addition, they are also very time-consuming. For dietary records, participants are asked to record details of the meals they consumed. Although this approach is less reliant on the participants' memory, it requires motivated and trained participants to accurately report their diet~\cite{Shim2014}. Another issue that affects the accuracy of these methods is that of under-reporting due to incorrect estimation of food portion sizes. Under-reporting has also been associated with factors such as obesity, gender, social desirability, restrained eating and hunger, education, literacy, perceived health status, age, and race/ethnicity~\cite{Zhu2010}. Therefore, there is an urgent need to develop new dietary assessment methods that can overcome these limitations. 

In the past decade, experts from the nutrition and engineering field have combined forces to develop new dietary assessment methods by leveraging technologies such as the Internet and mobile phones. Among the various new approaches, some of them use images captured at the eating scene to extract dietary information. These are called image-based dietary assessment methods. Examples of such methods include  TADA\texttrademark~\cite{Zhu2010}, FoodLog~\cite{food_log} , FoodCam~\cite{foodcam}, Snap-n-Eat~\cite{snap_eat}, GoCARB~\cite{gocarb}, DietCam~\cite{diet_cam} and ~\cite{mccrory2019methodology}, to name a few. In these methods, participants are asked to capture images of foods and drinks consumed via a mobile phone. These images are then analyzed to estimate the nutrient content. Estimating the nutrient content of foods in an image is commonly performed by trained dietitians, which can be time consuming, costly and laborious. More recently, automated methods have been developed to extract nutrient information of the foods from images~\cite{fang-globalsip2017, fang-ism2015, fang-icip2018}. The process of extracting nutrient information from images generally involves three sub-tasks,  food segmentation, food classification and portion size estimation~\cite{Zhu2010}.  Food image segmentation is the task of grouping pixels in an image representing foods. Food classification can then identify the food types. Portion size estimation~\cite{fang-ism2015} is the task of estimating the volume/energy of the foods in the image. Each of these tasks is essential for building an automated system to accurately extract nutrient information from food in images. In this paper, we focus on the task of food segmentation. In particular, we propose a food segmentation method that does not require information of the food types.

Food segmentation plays a crucial role in estimating nutrient information as the image segmentation masks are often used to estimate food portion sizes~\cite{diet_cam, fang-globalsip2017, fang-ism2015, food_portion1, food_portion2}. Food segmentation from a single image is a challenging problem as there is a large inter- and intra-class variance among different food types. Because of this variation, techniques developed for segmenting a particular class of foods will not be effective on other food classes. Despite these drawbacks, several learning based food segmentation methods~\cite{zhu-jbhi2015,Wang2017,yanai_food_seg, dehais2016food} have been proposed in recent years. One of the constraints of learning based methods is data dependency. They are only effective on the food categories they trained on.  For instance in~\cite{Wang2017}, class activation maps are used to segment food images. The Food-101 dataset~\cite{food-101} is used to train the model and the method is tested on a subset of another dataset that have common food categories with Food-101. This is a clear indication that their method~\cite{Wang2017} is only effective on food classes that have been trained on. Similarly, the learning based method proposed in~\cite{yanai_food_seg} is trained and tested only on UEC-FOOD100~\cite{uecfood-101}. The UEC-FOOD100 dataset has a total of 12,740 images with 100 different food categories, out of which 1,174 have multiple foods in a single image. In their method, the dataset is partitioned into training and testing subsets, each contains all the food categories. The authors of \cite{yanai_food_seg} split this dataset into training and testing in the following way. All the images containing a single food category were used for training and images containing multiple food categories were used for testing. This way the training set contained 11,566 images and the testing set contains 1,174 images. Splitting the dataset in this fashion does not guarantee that the training and testing subsets contain images belonging to different food categories. In fact this would mean they contain common food categories. Furthermore, the authors in \cite{yanai_food_seg} did not conduct any cross dataset evaluation. Thus the learning based method in \cite{yanai_food_seg} is also only effective on food categories it has been trained on. In~\cite{dehais2016food}, a semi automatic method is proposed to segment foods. The authors of~\cite{dehais2016food} assume that foods are always present in a circular region. In addition, they assume information about the number of different food categories is known. The experiments are conducted on a dataset of 821 images. While they achieved promising results, the proposed approach is not designed for real world scenario as their assumptions may not hold. In~\cite{chen2015saliency}, a food segmentation technique is proposed that exploits saliency information. However, this approach relies on successfully detecting the food container. In~\cite{chen2015saliency}, the food container is assumed to be a circular plate. Experimental results were reported using a dataset consisting of only 60 images. While the assumptions in~\cite{chen2015saliency} are valid in some cases, it may not be true in many real life scenarios. 

In addition, there are also constraints imposed by the available datasets. Publicly available food image datasets such as UECFOOD-100~\cite{uecfood-101}, Food-101~\cite{food-101} and UECFOOD-256~\cite{food-256} are biased towards a particular cuisine and also do not provide pixel level labelling. Pixel level labelling is crucial because it forms the necessary ground truth for training and evaluating learning based food segmentation methods. To overcome the limitations posed by learning based methods and the availability of public datasets with ground truth information, we proposed to develop a food segmentation method that is class-agnostic. In particular, our class-agnostic food segmentation method uses information from two images, the before eating and after eating image to segment the foods consumed during the meal. 

\begin{figure}[t!]
	\subfloat[Before eating image.]{\includegraphics[scale=.15]{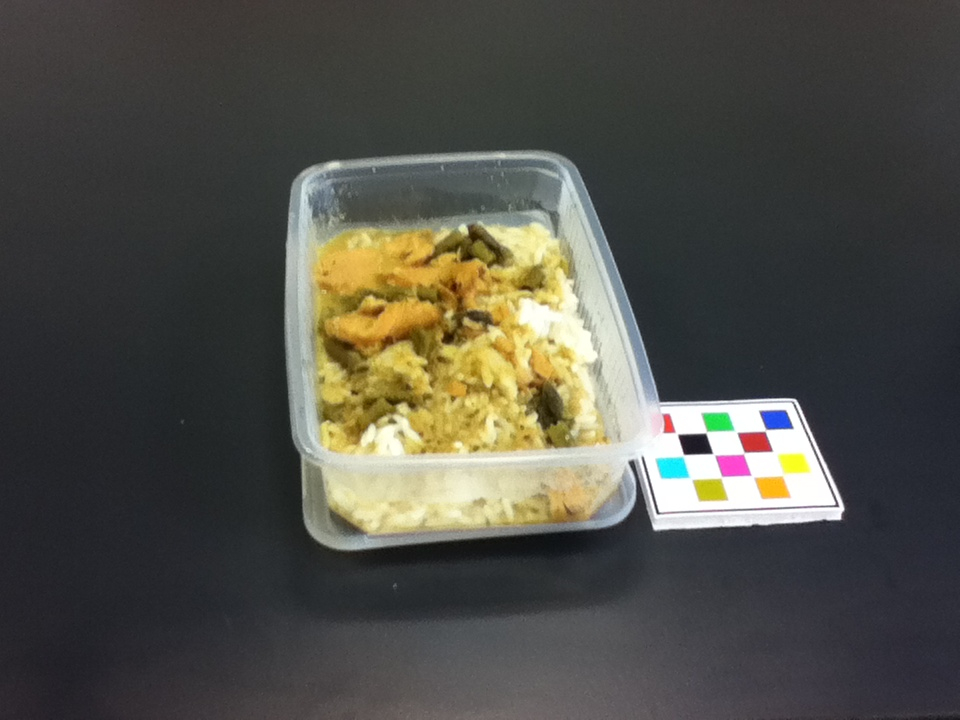}\label{fig:before_eating}}
	\hfill
	\subfloat[After eating image.]{\includegraphics[scale=.15]{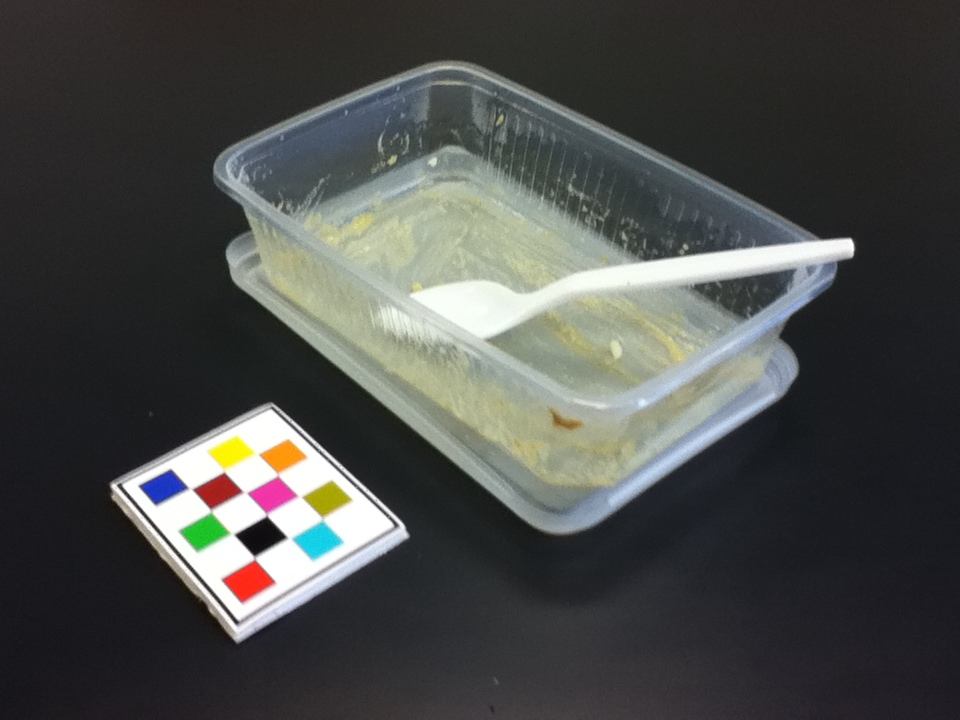}\label{fig:after_eating}}
	\caption{A pair of eating scene images, taken before and after a meal is consumed. The salient missing object in figure a is the food in the container.}
	\label{fig:example}
\end{figure}

Our data is collected from a community dwelling dietary study~\cite{tada4} using the TADA\texttrademark\ platform. In this study, participants were asked to take two pictures of their eating scene, one before they start eating which we call the before eating image and one immediately after they finished eating which we call the after eating image. The before eating and after eating image represent the same eating scene, however for the purpose of this work, we only select image pairs where the after eating image does not contain any food. Our goal is to segment the foods in the before eating image using information from both before and after eating images. To illustrated this problem in a more general scenario, lets consider an experimental setup in which a person is given a pair of images shown in Fig.\ref{fig:example} and is asked the following question, \enquote{Can you spot the salient objects in Fig.~\ref{fig:before_eating} that are missing in Fig.~\ref{fig:after_eating}?}. We refer to these as the \textit{salient missing objects}. To find salient missing objects, the Human Vision System (HVS) compares regions that are salient in both images. In this example, the food, container and color checkerboard in Fig. \ref{fig:before_eating} are the salient objects and in Fig. \ref{fig:after_eating}, the color checkerboard, spoon and container are the salient objects. Comparing the salient objects in both of these images, HVS can identify the food as the salient missing object. In this paper, our goal is to build a model to answer this question. By looking for salient missing objects in the before eating image using the after eating image as the reference we can then segment the foods without additional information such as the food classes. As the above approach does not require information about the food class, we are able to build a class-agnostic food segmentation method by segmenting only the salient missing objects. 

The above question does not bear significance for just any pair of random images. It only becomes relevant when the image pairs are related. For example, in Fig.~\ref{fig:example}, both images have many regions/objects with same semantic labels such as color checkerboard, container and the black background. However, the relative positions of these regions/objects are different in both images due to camera pose and different time of capturing the images. Because of similarity at the level of semantics between both images, it is plausible to define the notion of salient missing objects. Notice that we are not interested in pixel-level differences due to changes in illumination, poses and angles.

In this experimental scenario, the visual attention of HVS is guided via a task, hence it falls under the category of top down saliency. Visual attention ~\cite{6253254,6180177} is defined as the process that capacitates a biological or artificial vision system to identify relevant regions in a scene~\cite{6253254}. Relevance of every region in a scene is attributed through two different mechanisms, namely top down saliency and bottom up saliency. In top down saliency, attention is directed by a task. An example of this mechanism in action is how a human driver's HVS identifies relevant regions on the road for a safe journey. Other examples where top down saliency have been studied are sandwich making~\cite{BallardDanaH1995MRiN} and interactive game playing~\cite{game}. In bottom up saliency, attention is directed towards those regions that are the most conspicuous. Bottom up saliency is also known as visual saliency. In the real world, visual attention of HVS is guided by a combination of top down saliency and bottom up saliency. In the above question of finding salient missing objects, visual attention is guided by a task and hence it falls under the category of top down saliency. Top down saliency has not been studied as extensively as visual saliency because of its complexity~\cite{6253254}.

In this paper, we propose an unsupervised method to find the salient missing objects between a pair of images for the purpose of designing a class agnostic food segmentation method. We use the after eating image as the background to find the contrast of every pixel in the before eating image. We then fuse the contrast map along with saliency maps to obtain the final segmentation mask of the salient missing objects in the before eating image. We also compare our method to other class-agnostic methods. Since food is a salient object in the before eating image, by detecting salient objects in the before eating image we are able to segment the food. We compared our method to four state-of-the-art salient object detection methods, namely R3NET~\cite{r3net}, NLDF~\cite{nldf}, UCF~\cite{ucf} and Amulet~\cite{amulet}.

The paper is organized as follows. In Section~\ref{sec:related}, we formulate our problem and discuss related work. We describe our proposed method in detail in Section~\ref{sec:method}. In Section~\ref{sec:results}, we discuss dataset and experiment design. In Section~\ref{sec:discussion}, we discuss experimental results and compare our method with other salient object detection methods. Conclusions are provided in Section~\ref{sec:conclusion}.

\section{Problem Formulation and Related Work}
\label{sec:related}
In this section, we first introduce common notations used throughout the paper. We then discuss related works on modeling top down saliency and change detection.

\subsection{Problem Formulation}
Consider a pair of images $\{I^b, I^a\}$ captured from an eating scene.  

\begin{itemize}
	\item $I^b$ :  We refer to it as the \enquote{before eating image.}
	This is the meal image captured before consumption.
	\item $I^a$ : We refer to it as the \enquote{after eating image.} 
	This is the meal image captured immediately after consumption.
\end{itemize}
Our goal is to obtain a binary mask $M^b$, that labels the salient missing objects in $I^b$ as foreground (with a binary label of 1) and rest of $I^b$ as background (with a binary label of 0).

\subsection{Related Work}
Our goal is to find salient missing objects in a pair of images. Since the visual attention of the HVS is guided by a task, it falls under the category of top down saliency. Top down saliency is much more complex than visual saliency and hence has not been studied extensively. Some of the recent works modeling top down saliency paradigms are~\cite{7410695,DBLP:journals/corr/RamanishkaDZS16}. In~\cite{DBLP:journals/corr/RamanishkaDZS16}, given an image or video and an associated caption, the authors proposed a model to selectively highlight different regions  based on words in the caption. 
Our work is related in the sense that we also try to highlight and segment objects/regions based on a description, except that the description in our case is a much more generic question of finding the salient missing objects in a pair of images without specific details.   

Another related problem is modeling change detection~\cite{Khan2017LearningDS,BMVC2015_61,7948741,rengarajan:2014:efficient}. In change detection, the objective is to detect all relevant changes between a pair of images that are aligned or can be potentially aligned via image registration. Examples of such changes may include object motion, missing objects, structural changes~\cite{BMVC2015_61} and changes in vegetation~\cite{Khan2017LearningDS}. One of the key differences between change detection and our proposed problem is that in change detection, the pair of images are aligned or can be potentially aligned via image registration~\cite{szeliski2004image} which is not true in the case of salient missing objects. In the case of finding salient missing objects, we cannot guarantee that $I^b$ and $I^a$ can be registered, as often there is relative motion between objects of interest as shown in  Fig.~\ref{fig:example} and also in Fig.~\ref{fig:dataset}.

The problem of finding salient missing objects can be thought of as a change detection problem in a more complex environment than those that have been previously considered. Hence, we need to develop new methods to solve this problem.

\section{Method}
\label{sec:method}
\begin{figure*}[h]
\centering
\includegraphics[scale= .35]{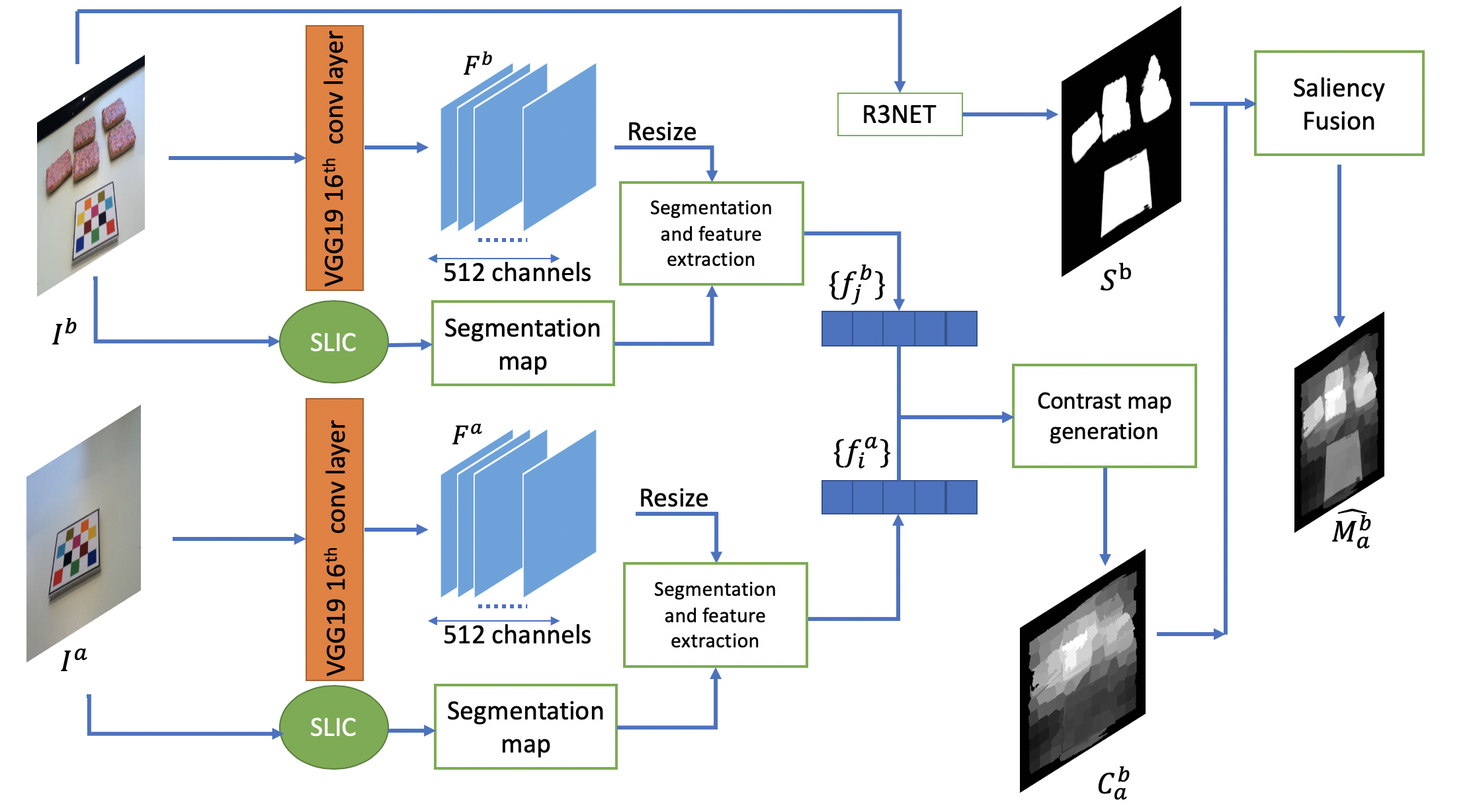}
\caption{Overview of proposed method.}
\label{fig:method}
\end{figure*}

In this section, we describe the details of our proposed method to segment salient missing objects in $I^b$. Our method consists of three parts, \textit{segmentation and feature extraction}, \textit{contrast map generation} and \textit{saliency fusion}. An overview of our proposed method is described in Fig.~\ref{fig:method}.

\subsection{Segmentation And Feature Extraction} 
\label{seg and feature}
We first segment the pair of images $I^a$ and $I^b$ using SLIC~\cite{slic} to group pixels into perceptually similar superpixels. Let $\mathcal{A} = \{a_i\}$ denote the superpixels of the after eating image $I^a$ and $\mathcal{B} = \{b_j\}$ for superpixels of the before eating image $I^b$. 

We extract features from each superpixel. We use these features to compute the contrast map. The contrast map $C_a^b$ gives an estimate of the probability of pixels belonging to objects/regions present in $I^b$ but missing in $I^a$. This will be explained in detail in section \ref{contrast}. To compute an accurate contrast map, pixels belonging to similar regions in $I^b$ and $I^a$ should have similar feature representation and vice versa. Going from $I^b$ to $I^a$ we can expect changes in  scene lightning,  changes in noise levels and changes in segmentation boundaries because of relative object motion. To compute an accurate contrast map, its important that feature representation of pixels are robust to these artifacts.For this reason, we extract features using a pretrained Convolutional Neural Network (CNN) instead of using hand-crafted features. We use the VGG19~\cite{VGG16} pretrained on the ImageNet dataset~\cite{imagenet_cvpr09}. ImageNet is a large dataset consisting of more than a million images belonging to 1000 different classes. It captures the distribution of natural images very well. Because of all these reasons models pretrained on ImageNet are widely used in several applications \cite{imagenet_1, imagenet_2, BMVC2015_61, sup2}.

We use the pretrained VGG19 for both $I^b$ and $I^a$. The output of $16^{th}$ convolutional layer in VGG19 is extracted as the feature map. The reasoning behind this choice is explained in section~\ref{hyper_sel}. According to Table 1 in \cite{VGG16}, VGG19 has a total of $16$ convolutional layers. The dimensionality of the output of the $16^{th}$ convolutinal layer of VGG19 is $14 \times 14 \times 512$ where $14 \times 14$ is the spatial resolution. The input ($I^b$ or $I^a$) to VGG19 has a spatial resolution of $224 \times 224$. We spatially upscale the output of the $16^{th}$ convolution layers by a factor of 16. 
We denote these upscaled feature maps of  $I^b$ and $I^a$ as $F^b$ and $F^a$, respectively. The dimensionality of $F^b$ and $F^a$ is then $224 \times 224 \times 512$. Thus every pixel will be represented by a 512 dimensional vector in the feature space. For each superpixel, we denote the extracted features as $\{f_j^b\}$ for the before eating image and $\{f_i^a\}$ for the after eating image. Using these extracted feature maps,  $f_i^a$ and $f_j^b$ are computed as described in Eq.~\ref{eq:eq1}.
\begin{equation} 
\label{eq:eq1}
\begin{split}
f_i^a = \frac{1}{m_i^a}\sum_{k\in r_i^a}F^a(k) \\
f_j^b = \frac{1}{m_j^b}\sum_{k\in r_j^b}F^b(k)
\end{split}
\end{equation}   
where $r_i^a$ denotes the set of pixels belongs to superpixel $a_i$ and $m_i^a$ is its cardinality. $r_j^b$ and $m_j^b$ are similarly defined. 

\subsection{Contrast Map Generation} \label{contrast}
Contrast is a term often associated with salient object detection methods. Contrast of a region in an image refers to its overall dissimilarity with other regions in the same image. It is generally assumed that regions with high contrast demand more visual attention~\cite{contrast}. In the context of our problem, visual attention is guided by trying to find objects in $I^b$ that are missing in $I^a$. Therefore, our contrast map $C^b_a$ of $I^b$ is an estimate of the probability of each pixel belonging to an object missing in $I^a$. $C^b_a$ is computed as shown in Eq.~\ref{eq2}.
\begin{equation} 
\label{eq2}
\begin{split}
C^b_a = \frac{C_{a}^{b,\text{local}} + C_{a}^{b,\text{neigh}}}{max(C_{a}^{b,\text{local}} + C_{a}^{b,\text{neigh}})} 
\end{split}
\end{equation}
In $C_{a}^{b,\text{local}}$, contrast values of a superpixel $b_j$ is computed using information from $b_j$ and $I^a$, while in $C_{a}^{b,\text{neigh}}$ contrast value of $b_j$ is computed using information from  $b_j$, its neighboring superpixels and $I^a$. $max(C_{a}^{b,\text{local}} + C_{a}^{b,\text{neigh}})$, which is the maximum value in the contrast map, is used to normalize $C^b_a$ to $[0, 1]$. To compute the contrast map $C_{a}^{b,\text{local}}$ or $C_{a}^{b,\text{neigh}}$, contrast values are computed for each superpixel and then these values are assigned to the associated individual pixels. However, if $b_j$ is a superpixel along the image boundaries, that is $b_j \in \mathcal{B}^e$, we assign $b_j$ a contrast value of zero. We assume that the salient missing objects are unlikely to be present along the image boundaries. 

The contrast value of a superpixel $b_j \notin \mathcal{B}^{e}$ is denoted by $c_j^{b, \text{local}}$, and is computed as:
\begin{equation} \label{eq3}
\begin{split}
c_j^{b, \text{local}} = \operatorname*{min}_{\forall i \ \text{such that} \  a_i \in \mathcal{A}} ||f_j^b - f_i^a||_2
\end{split}
\end{equation}
If $b_j \in \mathcal{B}^e$ then $c_j^{b, \text{local}} = 0$. $c_j^{b, \text{local}}$ is the minimum Euclidean distance between the feature vector $f_j^b$ and the closest feature vector of a superpixel in the after eating image. 
A superpixel $b_j$ belonging to objects/regions that are common to both $I^b$ and $I^a$ will have lower value of $c_j^{b, \text{local}}$, while $b_j$ belonging to objects/regions present in $I^b$ but missing in $I^a$ will likely have higher value of $c_j^{b, \text{local}}$. 

\begin{figure*}[ht]
	\centering
	\subfloat[$G_{a_{i_1}, b_{j_{25}}}$]{\includegraphics[scale= .34]{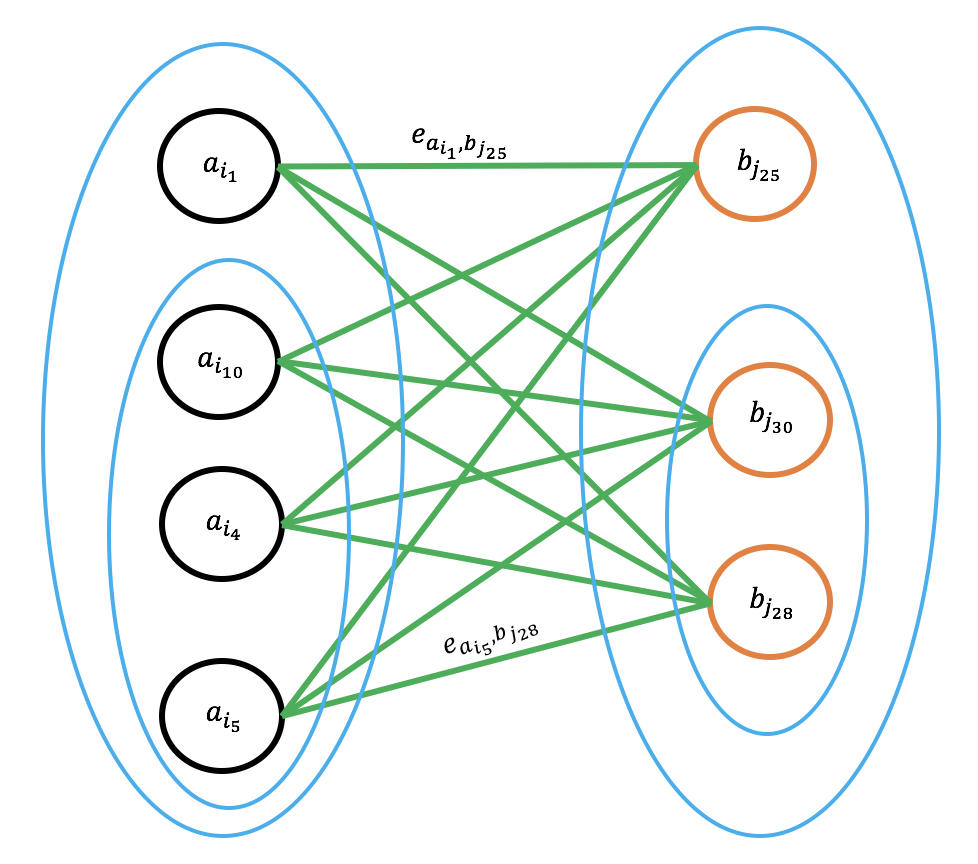}\label{fig:f1}}
	\hfill
	\subfloat[$\mathcal{S}^{\mathcal{E}_{a_{i_1}, b_{j_{25}}}}_{1} = \{ e_{a_{i_{4}}, b_{j_{28}}},e_{a_{i_{10}}, b_{j_{30}}}, e_{a_1, b_{j_{25}}} \}$]{\includegraphics[scale= .34]{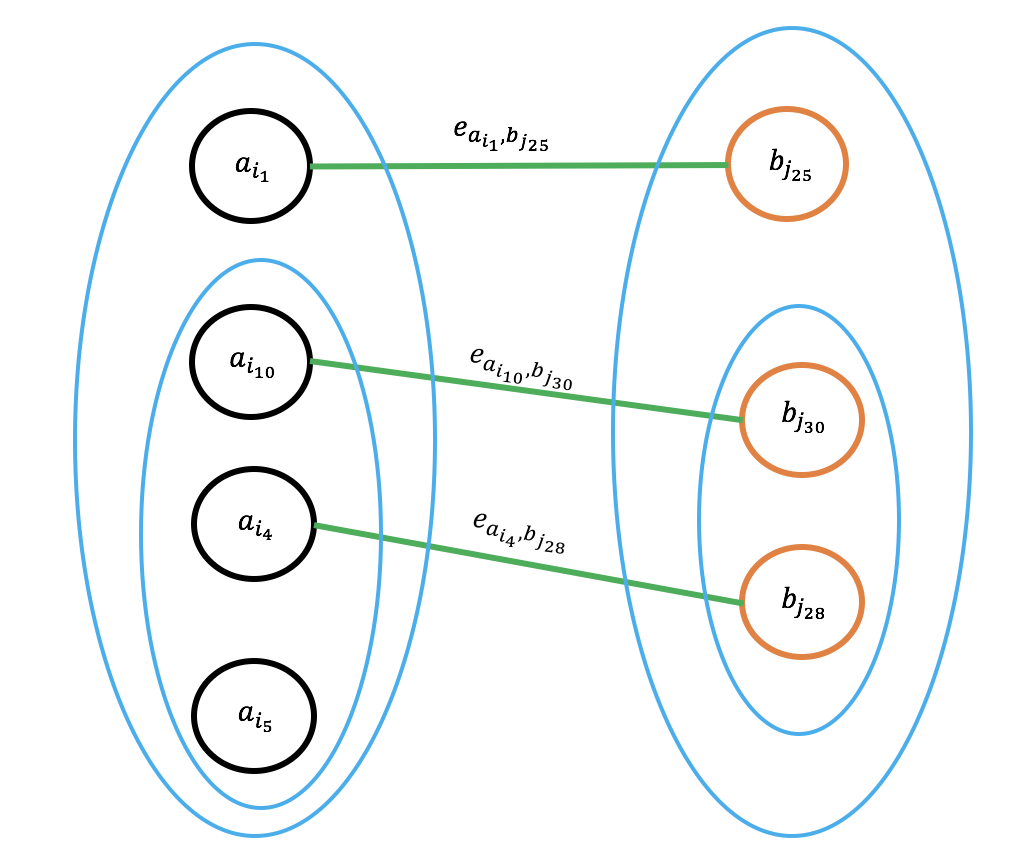}\label{fig:f2}}
	\hfill
	\subfloat[$\mathcal{S}^{\mathcal{E}_{a_{i_1}, b_{j_{25}}}}_{2} = \{e_{a_{i_{10}}, b_{j_{30}}},e_{a_{i_{4}}, b_{j_{28}}}, e_{a_5, b_{j_{25}}} \} $]{\includegraphics[scale= .36]{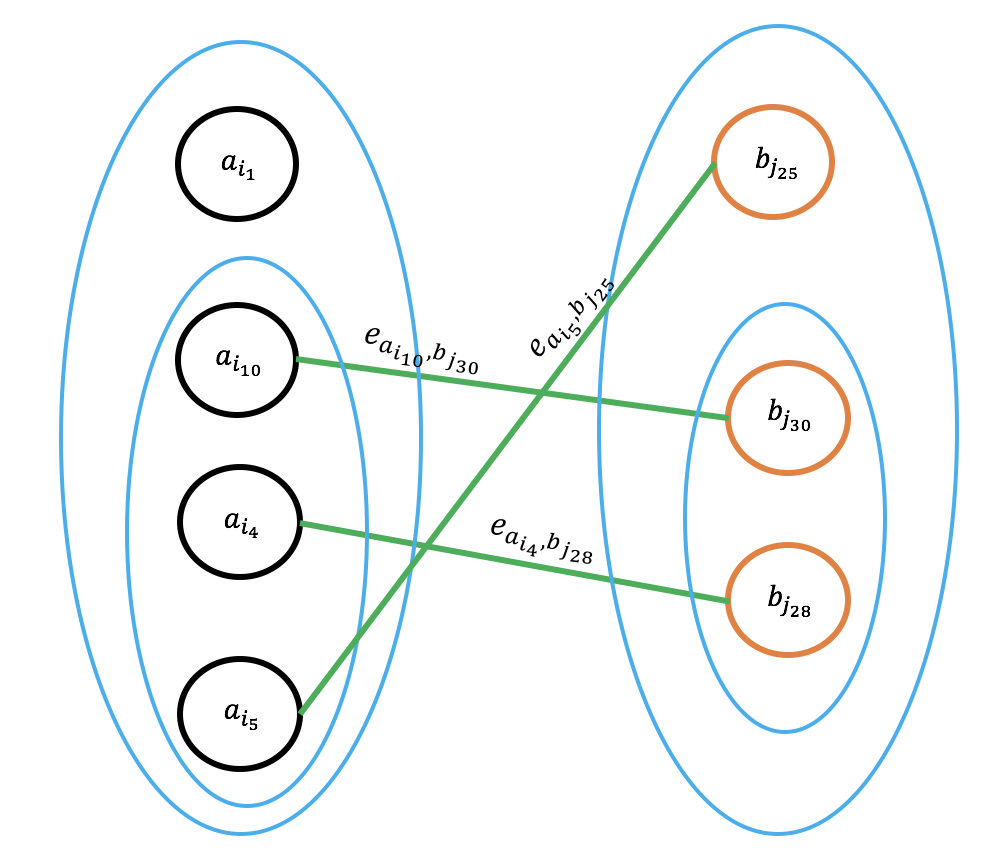}\label{fig:f3}}
	
	\caption{Consider 2 hypothetical nodes $a_{i_1} \in \mathcal{A}$ with $\mathcal{N}(a_{i_1}) = \{a_{i_{10}}, a_{i_{4}}, a_{i_{5}}\}$ and $b_{j_{25}} \in \mathcal{B}$  with $\mathcal{N}(b_{j_{25}}) = \{b_{j_{30}}, b_{j_{28}} \}$. In Fig.~\ref{fig:f1}, we illustrate how $G_{a_{i_1}, b_{j_{25}}}$ is constructed. Note that because $G_{a_{i_1}, b_{j_{25}}}$ is a complete bipartite graph there is an edge from every node in $\{a_{i_1}, \mathcal{N}(a_{i_1}) \}$ to every node in $\{b_{j_{25}}, \mathcal{N}(b_{j_{25}}) \}$. In Fig.~\ref{fig:f2} and Fig.~\ref{fig:f3}, examples of plausible maximum matching are shown. The value of $D(\mathcal{S}^{\mathcal{E}_{a_{i_1}, b_{j_{25}}}}_{1}) = w(e_{a_1, b_{j_{25}}}) + w(e_{a_{i_{10}}, b_{j_{30}}}) + w( e_{a_{i_{4}}, b_{j_{28}}} )$ and $D(\mathcal{S}^{\mathcal{E}_{a_{i_1}, b_{j_{25}}}}_{2})$ can be computed in a similar manner.}
	\label{fig:bgraph}
\end{figure*}

Before describing how we compute $C_{a}^{b,\text{neigh}}$, we need to introduce a few more notations. For a given superpixel $b_j$, let $\mathcal{N}(b_j)$ denote the set of all neighboring superpixels of $b_j$. Similarly, for any superpixel $a_i$, $\mathcal{N}(a_i)$ is the set of neighboring superpixels. Consider a complete bipartite graph over the two sets of superpixels $\{a_i, \mathcal{N}(a_i)\}$ and $\{b_j, \mathcal{N}(b_j)\}$ denoted by
\begin{equation} 
\label{eq4}
\begin{split}
{G_{{a_i},{b_j}}} = (\{ {a_i},{\mathcal{N}}({a_i})\}  \cup \{ {b_j},{\mathcal{N}}({b_j})\}, \ {{\mathcal{E}}_{{a_i},{b_j}}})
\end{split}
\end{equation}
where $\mathcal{E}_{a_i, b_j}$ is the set of edges in $G_{a_i, b_j}$. An example is shown in Fig.~\ref{fig:bgraph}.

In $G_{a_i, b_j}$, consider an edge $e_{a_{i_1}, b_{j_1}}$ between the two superpixels $a_{i_1} \in \ \{a_i, \mathcal{N}(a_i)\} $ and $b_{j_1} \in \ \{b_j, \mathcal{N}(b_j)\} $, the edge weight is evaluated by the Euclidean norm $w(\cdot)$ defined as:
\begin{equation}  \label{eq5}
\begin{split}
w(e_{a_{i_1}, b_{j_1}}) = ||f_{a_{i_1}} - f_{b_{j_1}}||_2
\end{split}
\end{equation}
 A matching over $G_{a_i, b_j}$ is a set of edges $\mathcal{S} \subset \mathcal{E}_{a_i, b_j}$ such that no two edges in $\mathcal{S}$  share the same nodes. A maximum matching over $G_{a_i, b_j}$, denoted by $\mathcal{S}^{\mathcal{E}_{a_i, b_j}}_{k} \subset \mathcal{E}_{a_i, b_j}$, is a matching of maximum cardinality. There can be many possible maximum matchings over $G_{a_i, b_j}$, hence we use subscript $k$ in $\mathcal{S}^{\mathcal{E}_{a_i, b_j}}_{k}$ to denote one such possibility. The cost of a given $\mathcal{S}^{\mathcal{E}_{a_i, b_j}}_{k}$ is denoted by $D(\mathcal{S}^{\mathcal{E}_{a_i, b_j}}_{k})$ and is defined as:
\begin{equation} \label{eq6}
\begin{split}
D({\mathcal{S}}_k^{{{\mathcal{E}}_{{a_i},{b_j}}}}) = \sum\limits_{\forall e \in S_k^{{\varepsilon _{{a_i},{b_j}}}}} {w(e)} 
\end{split}
\end{equation}
Given a $G_{a_i, b_j}$, we want to find the maximum matching with the minimum cost. We refer to this minimum cost as $\hat{D}_{\text{min}}(G_{a_i, b_j})$ and it is computed as:
\begin{equation} \label{eq7}
\begin{split}
\hat{D}_{\text{min}}(G_{a_i, b_j}) = \operatorname*{\text{min}}_{\forall k \ \text{such that} \  \exists \ \mathcal{S}^{\mathcal{E}_{a_i, b_j}}_{k}} D(\mathcal{S}^{\mathcal{E}_{a_i, b_j}}_{k})
\end{split}
\end{equation}
For two superpixels $a_i$ and $b_j$,  $\hat{D}_{\text{min}}(G_{a_i, b_j})$ measures the similarity between the two superpixels and the similarity between their neighborhoods. The lower the value of $\hat{D}_{\text{min}}(G_{a_i, b_j})$, the more similar the two superpixels are both in terms of their individual characteristics and their neighboring superpixels. The contrast value of superpixel $b_j \notin \mathcal{B}^{e}$ in $C_{a}^{b,\text{neigh}}$ is denoted by $c_j^{b, \text{neigh}}$  and is computed as:
\begin{equation} \label{eq8}
\begin{split}
c_j^{b, \text{neigh}} = \operatorname*{\text{min}}_{\forall i \ \text{such that} \ a_i \in \mathcal{A}}\frac{\hat{D}_{\text{min}}(G_{a_i, b_j})}{l_{a_i, b_j}}
\end{split}
\end{equation}
In Eq.~\ref{eq8}, $l_{a_i, b_j} = \operatorname*{\text{min}}({|\{a_i, \mathcal{N}(a_i)\}|, |\{b_j, \mathcal{N}(b_j)\}|})$ where $|\{.\}|$ denotes the cardinality of the set $\{.\}$. If $b_j \in \mathcal{B}^{e}$ then $c_j^{b, \text{neigh}} = 0$. $\hat{D}_{\text{min}}(G_{a_i, b_j})$ is likely to increase as $l_{a_i, b_j}$ increases because  there are more edges in maximum matching. In order to compensate this effect, we divide $\hat{D}_{\text{min}}(G_{a_i, b_j})$ by  $l_{a_i, b_j}$ in Eq.~\ref{eq8}. 

\subsection{Saliency Fusion}
\label{salfus}
The contrast map $C_a^b$ gives an estimate of the probability of pixels belonging to objects/regions present in $I^b$ but missing in $I^a$. However, we would like to segment salient missing objects. As explained in Section~\ref{intro}, to find the salient missing objects, the HVS compares objects/regions in $I^b$ that have a high value of visual saliency. Therefore, we are interested in identifying regions in the contrast map $C^b_a$ which correspond to high visual saliency. The visual saliency information of $I^b$ needs to be incorporated into $C^b_a$ to obtain our final estimate $\hat{M}_a^b$, where $\hat{M}_a^b$ is the probability of each pixel in $I^b$ belonging to the salient missing objects. We can then obtain the final binary label $M^b$, by thresholding $\hat{M}_a^b$ with $T \in [0, 1]$. If $S^b$ is the visual saliency map of $I^b$, then $\hat{M}_a^b$ is computed as:
\begin{equation} \label{eq9}
\begin{split}
\hat{M}_a^b = \frac{\alpha*S^b + C^b_a}{max(\alpha*S^b + C^b_a)}
\end{split}
\end{equation}
where $max(\alpha*S^b + C^b_a)$ is the normalization term. In Eq.~\ref{eq9}, $\alpha$ is a weighting factor between $[0, 1]$ that varies the relative contributions of $S^b$ and $C^b_a$ towards $\hat{M_a^b}$. The value of $\alpha$ is empirically computed and will be explained in Section~\ref{exp_ana}. To compute $S^b$, we use the state-of-the-art salient object detection method R3NET~\cite{r3net}. We also compared our method to other deep learning based salient object detection methods such as Amulet~\cite{amulet}, UCF~\cite{ucf} and NLDF~\cite{nldf}. 

\section{Experimental Results}
\label{sec:results}
\subsection{Dataset}
The dataset $\mathcal{D}$ we use for evaluating our method contains 566 pairs of before eating and after eating images. Along with image pairs, ground truth masks of the salient missing objects in the before eating images (which in this case are foods) are also provided. These images are a subset of images collected from a community dwelling dietary study~\cite{tada4}. 
The images in $\mathcal{D}$ exhibit a wide variety of foods and eating scenes.  
Participants in this dietary study are asked to capture a pair of before and after eating scene images, denoted as $I^b$ and $I^a$. A typical participant takes about 3 to 5 pairs of images per day depending on his/her eating habits. These image pairs are then sent to a cloud based server to analyze nutrient contents. $\mathcal{D}$ is split randomly into $\mathcal{D}_{val}$ (49 image pairs) and $\mathcal{D}_{test}$ (517 image pairs). $\mathcal{D}_{val}$  is used for choosing the optimal hypyerparameters namely $\alpha$ and the convolutional layer. More details are explained in section~\ref{hyper_sel}. $\mathcal{D}_{test}$  is used to evaluate the accuracy of our method compared to other methods. Examples of image pairs from $\mathcal{D}_{test}$ along with the predicted masks obtained by our method and the salient object detection methods are shown in Fig.~\ref{fig:dataset}. $\mathcal{D}_{test}$ and $\mathcal{D}_{val}$ have very different food classes. In addition, the background of the images in $\mathcal{D}_{val}$ is very different from those in $\mathcal{D}_{test}$. This makes $\mathcal{D}$ very apt for our experiments, because  $\mathcal{D}_{val}$ does not give any information about the food classes present in  $\mathcal{D}_{test}$. Thus if a model tuned on  $\mathcal{D}_{val}$  performs well on  $\mathcal{D}_{test}$, it signifies that the model is able to segment foods without requiring information about the food class.

\subsection{Evaluation Metrics}
We use two standard metrics for evaluating the performance of the proposed method. These metrics are commonly used to assess the quality of salient object detection methods~\cite{survey}.  
\begin{itemize}
	\item \textbf{Precision and Recall} Consider $t = \{I^b, I^a, G^b\}$ in $\mathcal{D}$. In $t$, $G^b$ represents the ground truth mask of the salient missing objects in $I^b$. Pixels belonging to the salient missing objects in $G^b$ have a value of $1$ and the rest have a value of $0$. Our proposed method outputs $\hat{M}^b_a$ which has a range between $[0,1]$. We can then generate a segmentation mask $M^b$ using a threshold $T \in [0, 1]$. Given $M^b$ and $G^b$, precision (P) and recall (R) are computed over $\mathcal{D}$ as:
	\begin{equation} \label{eq10}
	\begin{split}
	\text{P \ :} \ \frac{\operatorname*{\sum}_{\forall t \in \mathcal{D}}|M^b \cap G^b|}{\operatorname*{\sum}_{\forall t \in \mathcal{D}}|M^b|} \ , \ \text{R \ :} \ \frac{\operatorname{\sum}_{\forall t \in \mathcal{D}}|M^b \cap G^b|}{\operatorname{\sum}_{\forall t \in \mathcal{D}}|G^b|}
	\end{split}
	\end{equation}
	For a binary mask, $|\cdot|$ denotes the number of non-zero entries in it. By varying $T$ between 0 and 1, we have different pairs of precision and recall values. When precision and recall values are plotted against each other, we obtain the precision recall (PR) curve. The information provided by precision and recall can be condensed into their weighted harmonic mean denoted by $F_{\beta}$, where $F_{\beta}$ is computed as: 
	\begin{equation} \label{eq11}
	\begin{split}
	F_{\beta} = \frac{(1+\beta^2)*Precision*Recall}{\beta^2*Precision + Recall}
	\end{split}
	\end{equation}
	The value of  lies between $[0, 1]$.  A higher values of $F_{\beta}$ indicates better performance. 
	The value of $\beta^2$ is chosen to be 0.3 similar to other works~\cite{survey}. $\beta$ is a control parameter that emphasizes the importance of precision over recall. The value $F_\beta$ varies as we move along the PR curve. The entire information of PR curve can be summarized by the maximal $F_\beta$ denoted by $F_\beta^{\text{max}}$, as discussed in~\cite{survey, fbeta}. 
	
	\item \textbf{Receiver Operator Characteristics (ROC)} Similar to the PR curve,  ROC curve is a plot of the true positive rate (TPR) against the false positive rate (FPR). TPR and FPR are defined as:
	 \begin{equation} \label{eq12}
	 \begin{split}
	 \text{TPR:} \; \frac{\operatorname{\sum}_{\forall t \in \mathcal{D}}|M^b \cap G^b|}{\operatorname{\sum}_{\forall t \in \mathcal{D}}|G^b|} \ , \ 
	 \text{FPR:} \; \frac{\operatorname{\sum}_{\forall t \in \mathcal{D}}|M^b \cap (1-G^b)|}{\operatorname{\sum}_{\forall t \in \mathcal{D}}|(1-G^b)|}
	 \end{split}
	 \end{equation}

Similar to $F_\beta^{max}$, the entire information provided by ROC curve can be condensed into one metric called AUC, which is the area under the ROC curve. Higher values of AUC indicate better performance. A perfect method will have an AUC of 1 and a method that randomly guesses values in $M^b$ will have an AUC of 0.5. 

\end{itemize}

\begin{figure}[t!]
	\subfloat[ VGG19 $F_\beta^{\text{max}}$ vs $\alpha$]{\includegraphics[scale= .4]{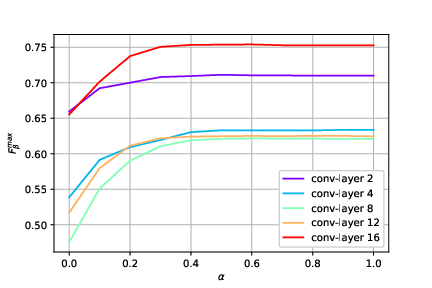}\label{fig:fbeta_alpha}}
	\hfill
	\subfloat[ ResNet34 $F_\beta^{\text{max}}$ vs $\alpha$]{\includegraphics[scale= .4]{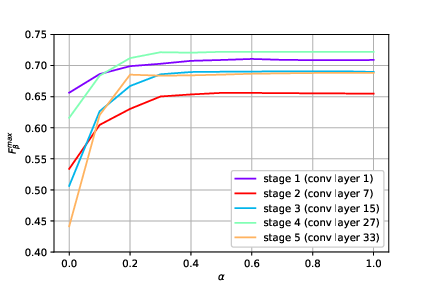}\label{fig:fbeta_alpha_resnet}}
	\hfill
	\subfloat[ Inception-v3 $F_\beta^{\text{max}}$ vs $\alpha$]{\includegraphics[scale= .4]{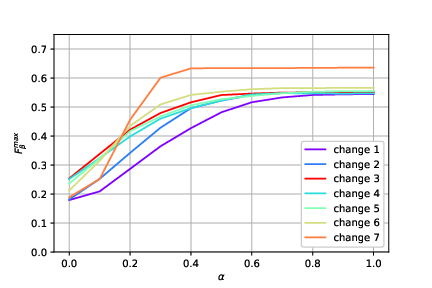}\label{fig:fbeta_alpha_inception}}
	\caption{ $F_\beta^{\text{max}}$ of $\hat{M}^b_a$ on $\mathcal{D}_{val}$ are plotted as $\alpha$ varies. (a) For VGG19, $F_\beta^{\text{max}}$ is reported using features from all convolutional layers that precede a max polling layer. (b) For ResNet34, features were extracted from the output of each stage. (c) For Inception-V3, features were extracted from each layer whenever the output spatial dimensions do not match the input spatial dimensions.}
	\label{fig:alpha}
\end{figure}

\begin{figure}
\centering
\subfloat[ROC curve]{\includegraphics[height = 6cm, width =8cm]{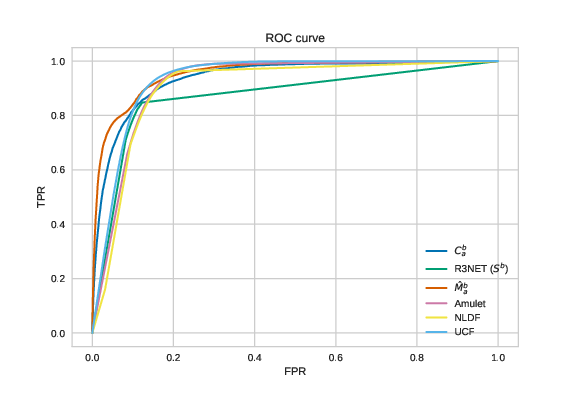}\label{fig:roc}}
\hfill
\subfloat[ROC curve (Zoomed)]{\includegraphics[height = 7cm, width =6cm]{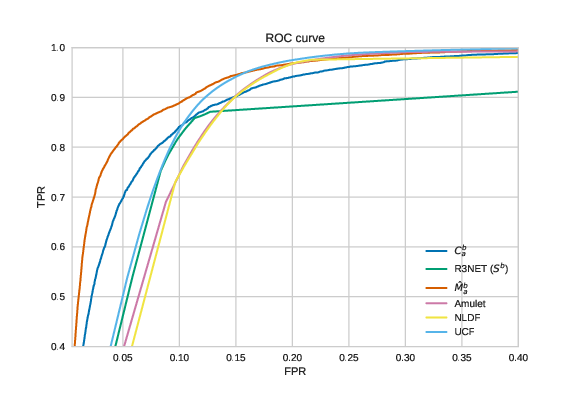}\label{fig:roc_foc}}
\hfill
\subfloat[PR curve]{\includegraphics[height = 6cm, width = 10cm]{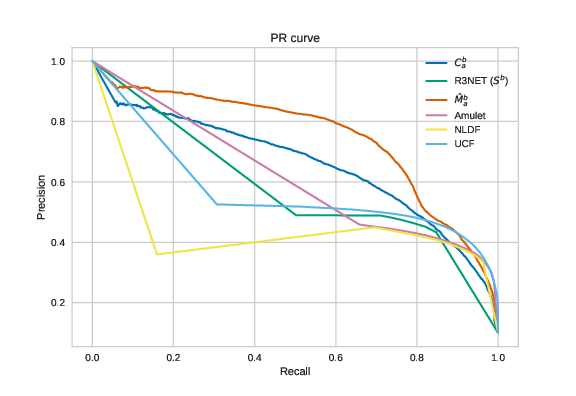}\label{fig:pr}}
\caption{ROC  and PR  curves of  R3NET~\cite{r3net} (also $S^b$), NLDF~\cite{nldf}, Amulet~\cite{amulet}, UCF~\cite{ucf}, $C^b_a$ and  $\hat{M}^b_a$ are shown in the above plots. Fig~\ref{fig:roc_foc} is a zoomed in version of ROC curve in Fig~\ref{fig:roc}}
\label{fig:pr_roc}
\end{figure}

\subsection{Experiments} 
\label{exp_ana}
\subsubsection{Hyperparameter selection}
\label{hyper_sel}
The method described in Section 3 requires 2 hyperparameters, namely $\alpha$ in Eq. \ref{eq9} and the convolutional layer of VGG19 for feature extraction. To justify the use of a pre-trained VGG19 for feature extraction, we have also conducted experiments by extracting features from  ResNet34~\cite{resnet34} and Inception-v3~\cite{inception}, pre-trained on ImageNet. These experiments are conducted on $\mathcal{D}_{val}$ to find the best $F_{\beta}$ which gives us a set of optimal hyperparameters.

To choose the best convolutional layer, we evaluate $\hat{M}^b_a$ using features from every convolutional layer of VGG19 that precedes a max pooling layer. There are 5 such convolutional layers in VGG19. The architecture of ResNet34 can be divided into 5 stages \cite{resnet34}. To find the optimal layer in ResNet34, we extracted features from the output of each stage. The architecture of Inception-v3 is very different from those of ResNet34 and VGG19. To find the optimal layer in Inception-v3, we extract features whenever there is a change in spatial dimension as the inputs propagate through the network. There are 7 such changes occur in Inception-v3 before the average pooling operation. Please refer to architecture of Inception-v3 provided in PyTorch \cite{pytorch} for more details. In addition to extracting features from various convolutional layers, we also vary $\alpha$ from $0$ to $1$ in steps of $0.1$. We plot $F_{\beta}^{max}$ as  $\alpha$ varies for every convolutional layer. The result is shown in Fig.~\ref{fig:alpha}. From Fig.~\ref{fig:alpha}, its quite evident that features from the $16^{th}$ convolutional layer gives the best performance compared to features from other layers. In addition it's also evident that features from VGG19 achieve better performance than features from ResNet34. For features from VGG19, the value of $F_{\beta}^{max}$ attains its maximum value of $0.754$ for $\alpha = 0.6$.

As we go deeper into the convolutional layers of VGG19, the features extracted become increasingly abstract, but suffer from decrease in resolution. Abstract features are less prone to changes in illumination, noise and pose which suits our task well. We noticed in Figure~\ref{fig:alpha}, as we go deeper into the convolutional layers, we first observe a degradation in the quality of features extracted (conv-layer 2 to conv-layer 8). This trend is reversed from conv-layer 8 to conv-layer 16 with a significant improvement of $F_{\beta}^{\text{max}}$. We suspect this is because at first the negative effect of decreased resolution outweighs the benefit of abstract features. However, this trend quickly reverses from conv-layer 8 and beyond.

\subsubsection{Testing}
After obtaining the optimal hyperparameters as described in section \ref{hyper_sel}, we evaluated our method on $\mathcal{D}_{test}$. $\hat{M}^b_a$ is computed for every image pair in $\mathcal{D}_{test}$ and the ROC and PR curves are computed on $\mathcal{D}_{test}$. Since our goal is to develop a class-agnostic food segmentation method, we compared the proposed method to 4 state-of-the-art salient object detection techniques, namely R3NET~\cite{r3net}, NLDF~\cite{nldf}, 
Amulet~\cite{amulet} and UCF~\cite{ucf}. Salient object detection methods  are class-agnostic and are applicable in this scenario as food is always a salient object in $I_b$. Since these are deep learning based methods, we use their respective pre-trained models to compute the saliency maps of $I_b$. The ROC and PR curves of various methods are shown in Fig.~\ref{fig:pr_roc}. The $F_{\beta}^{max}$ and AUC values are reported in Table~\ref{tb:auc_f}.

\begin{center}
\captionof{table}{AUC and $F_\beta^{\text{max}}$ values of various maps and methods.}
\begin{tabular}{ccc}
	\toprule
	Maps & AUC & $F_\beta^{\text{max}}$	 \\
	\midrule
	$C^{b}_a$ & 0.937 & 0.645 \\
	R3NET~\cite{r3net} ($S^b$) & 0.871 & 0.527 \\
	$\mathbf{\hat{M}^b_a}$ (ours) & \textbf{0.954} & \textbf{0.741} \\
    Amulet~\cite{amulet} & 0.919 &  0.499 \\
	NLDF~\cite{nldf} & 0.909 & 0.493 \\
	UCF~\cite{ucf} & 0.934 & 0.536   \\
    \bottomrule
\end{tabular}
\label{tb:auc_f}
\end{center}

\section{Discussion}
\label{sec:discussion}

\begin{figure*}[t!]
\captionsetup[subfigure]{labelformat=empty}
	\centering
	\hfill
	\subfloat{\includegraphics[scale = .2]{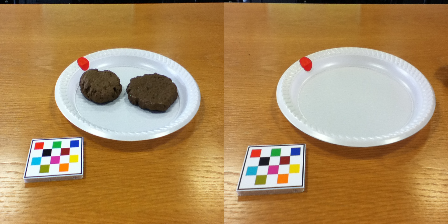}}
	\hfill
	\subfloat{\includegraphics[scale = .2]{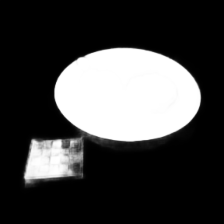}}
	\hfill
	\subfloat{\includegraphics[scale = .2]{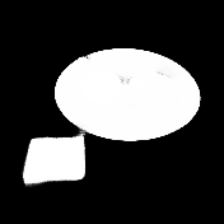}}
	\hfill
	\subfloat{\includegraphics[scale = .2]{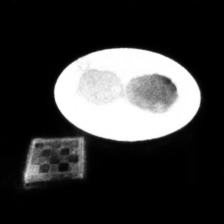}}
	\hfill
	\subfloat{\includegraphics[scale = .2]{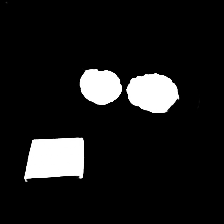}}
	\hfill
	\subfloat{\includegraphics[scale = .2]{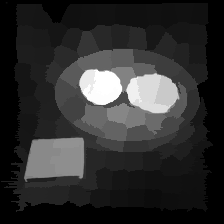}}
	\hfill
	\subfloat{\includegraphics[scale = .2]{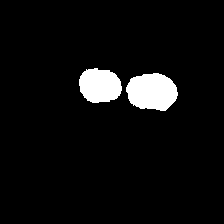}}
	
	\hfill
	\subfloat{\includegraphics[scale = .2]{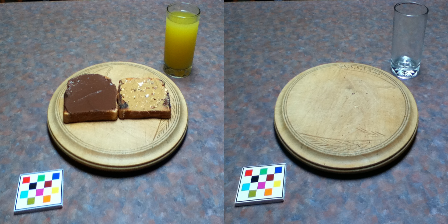}}
	\hfill
	\subfloat{\includegraphics[scale = .2]{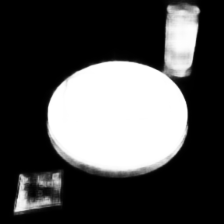}}
	\hfill
	\subfloat{\includegraphics[scale = .2]{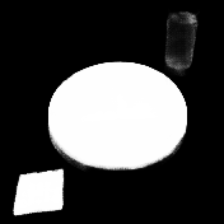}}
	\hfill
	\subfloat{\includegraphics[scale = .2]{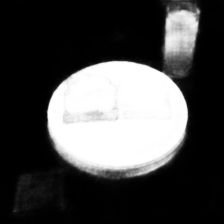}}
	\hfill
	\subfloat{\includegraphics[scale = .2]{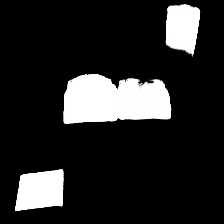}}
	\hfill
	\subfloat{\includegraphics[scale = .2]{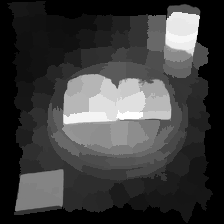}}
	\hfill
	\subfloat{\includegraphics[scale = .2]{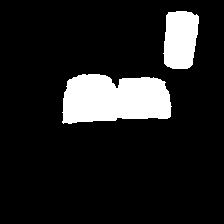}}

	\hfill
	\subfloat{\includegraphics[scale = .2]{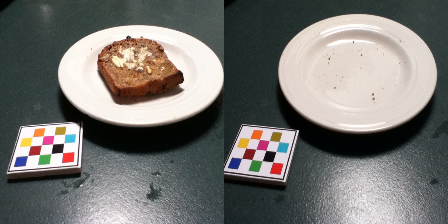}}
	\hfill
	\subfloat{\includegraphics[scale = .2]{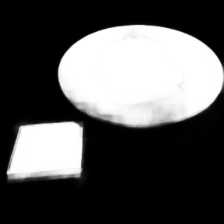}}
	\hfill
	\subfloat{\includegraphics[scale = .2]{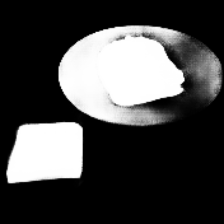}}
	\hfill
	\subfloat{\includegraphics[scale = .2]{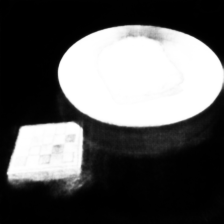}}
	\hfill
	\subfloat{\includegraphics[scale = .2]{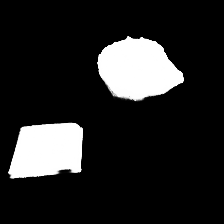}}
	\hfill
	\subfloat{\includegraphics[scale = .2]{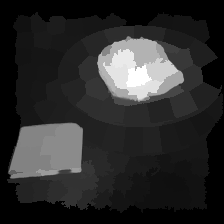}}
	\hfill
	\subfloat{\includegraphics[scale = .2]{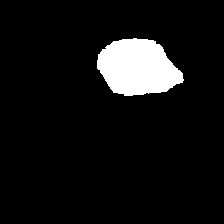}}

    \hfill
	\subfloat{\includegraphics[scale = .2]{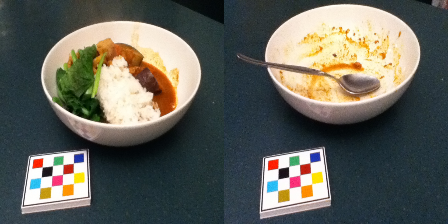}}
	\hfill
	\subfloat{\includegraphics[scale = .2]{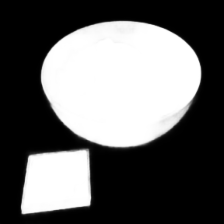}}
	\hfill
	\subfloat{\includegraphics[scale = .2]{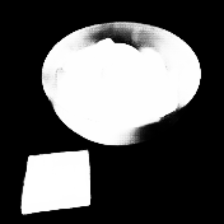}}
	\hfill
	\subfloat{\includegraphics[scale = .2]{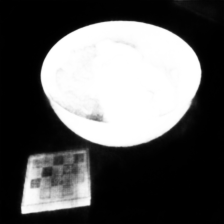}}
	\hfill
	\subfloat{\includegraphics[scale = .2]{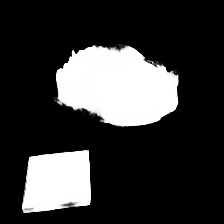}}
	\hfill
	\subfloat{\includegraphics[scale = .2]{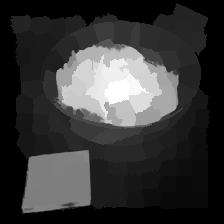}}
	\hfill
	\subfloat{\includegraphics[scale = .2]{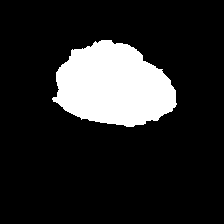}}
    
    \hfill
	\subfloat{\includegraphics[scale = .2]{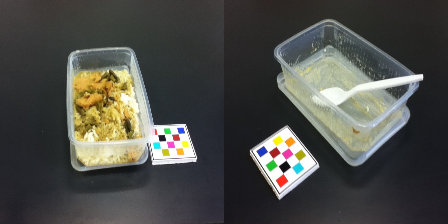}}
	\hfill
	\subfloat{\includegraphics[scale = .2]{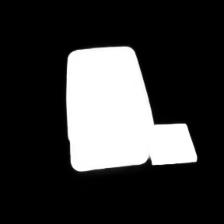}}
	\hfill
	\subfloat{\includegraphics[scale = .2]{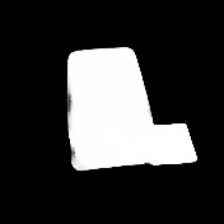}}
	\hfill
	\subfloat{\includegraphics[scale = .2]{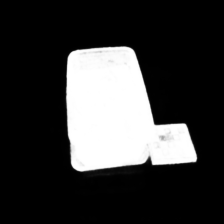}}
	\hfill
	\subfloat{\includegraphics[scale = .2]{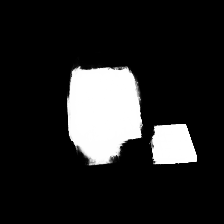}}
	\hfill
	\subfloat{\includegraphics[scale = .2]{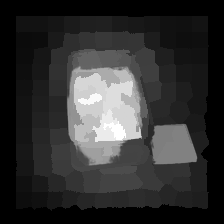}}
	\hfill
	\subfloat{\includegraphics[scale = .2]{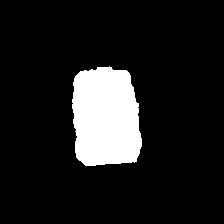}}

    \hfill
	\subfloat{\includegraphics[scale = .2]{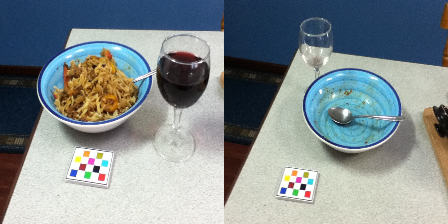}}
	\hfill
	\subfloat{\includegraphics[scale = .2]{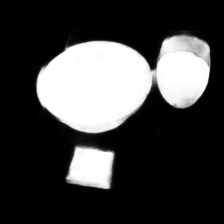}}
	\hfill
	\subfloat{\includegraphics[scale = .2]{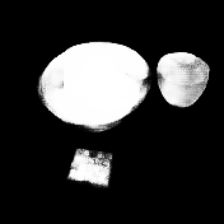}}
	\hfill
	\subfloat{\includegraphics[scale = .2]{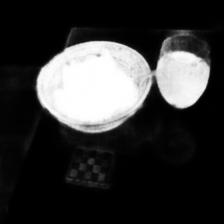}}
	\hfill
	\subfloat{\includegraphics[scale = .2]{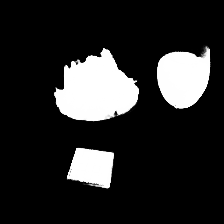}}
	\hfill
	\subfloat{\includegraphics[scale = .2]{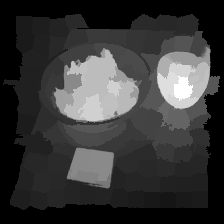}}
	\hfill
	\subfloat{\includegraphics[scale = .2]{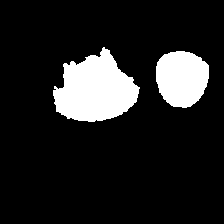}}

    \hfill
	\subfloat[Image Pairs.]{\includegraphics[scale = .2]{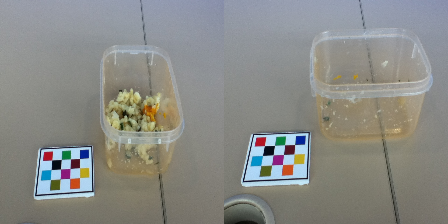}}
	\hfill
	\subfloat[Amulet\cite{amulet}]{\includegraphics[scale = .2]{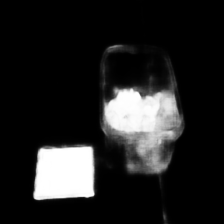}}
	\hfill
	\subfloat[UCF\cite{ucf}]{\includegraphics[scale = .2]{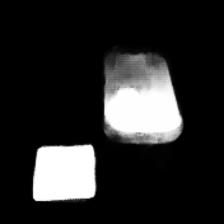}}
	\hfill
	\subfloat[NLDF\cite{nldf}]{\includegraphics[scale = .2]{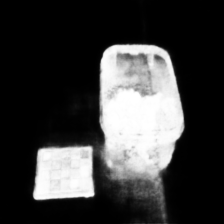}}
	\hfill
	\subfloat[R3NET\cite{r3net}]{\includegraphics[scale = .2]{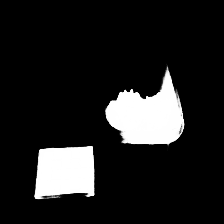}}
	\hfill
	\subfloat[Ours $\hat{M}^b_a$]{\includegraphics[scale = .2]{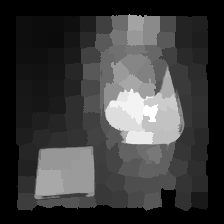}}
	\hfill
	\subfloat[Ground Truth $G^b$]{\includegraphics[scale = .2]{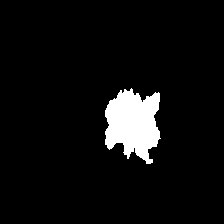}}
	\caption{Sample image pairs from $\mathcal{D}_{test}$ along with various maps are shown. For every row, the first group of two images are the original before and after eating images, respectively. The second group of images are the saliency maps generated by Amulet~\cite{amulet}, UCF~\cite{ucf}, NLDF~\cite{nldf}, R3NET~\cite{r3net} ,  $M^b_a$ (our method) followed by ground truth mask $G^b$. The ground truth images are binary maps with pixels of value 1 representing foods and pixels of value 0 representing background. All the others are probability maps with pixels having values between 0 and 1.}
	\label{fig:dataset}
	
\end{figure*}

The goal of our method is to  segment the salient missing objects in $I^b$ using information from a pair of images $I^a$ and $I^b$. In the contrast map generation step as described in Section~\ref{contrast}, we provide an estimate of the probability of pixels belonging to objects/regions in $I^b$ but missing in $I^a$. In the saliency fusion step as described in Section~\ref{salfus}, saliency information of pixels in $I^b$ is fused into the contrast map $C^b_a$ so as to emphasize that we are looking for salient missing objects. In order to show that the various steps of our proposed method achieve their individual objectives, we plotted the PR and ROC curves of the contrast map $C^b_a$, the visual saliency map $S^b$ from R3NET~\cite{r3net} and the estimated salient missing objects probability map $\hat{M}^b_a$ in Fig.~\ref{fig:pr} and Fig.~\ref{fig:roc}. In addition, we also plot PR and ROC curves for the 3 other salient object detection methods.  From these plots, we can see that combining $S^b$ and $C^b_a$ as described in Section~\ref{salfus} improves the overall performance. This is also illustrated in Table~\ref{tb:auc_f}, where both AUC and $F_\beta^{\text{max}}$  of $\hat{M}^b_a$  are higher than $C^b_a$. This is because the contrast map $C^b_a$ by itself models all the missing objects/regions while the probability map $\hat{M}^b_a$ also takes into account the visual saliency map $S^b$, which can more accurately model the salient missing objects. We can also observe from the PR and ROC curves in Fig.~\ref{fig:pr_roc} and values in Table\ref{tb:auc_f} that our method achieved better performance than the state-of-the-art salient object detection methods such as R3NET~\cite{r3net}, NLDF~\cite{nldf}, Amulet~\cite{amulet} and UCF~\cite{ucf}. We also visually verify the performance of our method as illustrated in Fig.~\ref{fig:dataset}. The salient object detection methods Amulet~\cite{amulet}, UCF~\cite{ucf} and NLDF~\cite{nldf} failed to detect only foods in these images, while R3NET~\cite{r3net} succeeded in detecting the foods but also placed equal importance to other salient objects such as the color checkerboard. Our method gave higher probability to the foods which are the salient missing objects compared to other salient objects in the scene. It must also be noted that our method did not have access to information about food classes in $\mathcal{D}_{test}$. This is because  $\mathcal{D}_{val}$ and $\mathcal{D}_{test}$ have very few food classes in common. By tuning the parameters on $\mathcal{D}_{val}$, our method will not have access to information about the food classes in  $\mathcal{D}_{test}$. Hence the performance of our method on $\mathcal{D}_{test}$ is indicative of its effectiveness of segmenting foods in a class-agnostic manner. These unique characteristics of $\mathcal{D}$ are also explained in section 4.1. Hence, by modeling the foods as salient missing objects, we are able to build a better class-agnostic food segmentation method compared to existing methods.

\section{Conclusion}
\label{sec:conclusion}
In this paper, we propose a class-agnostic food segmentation method by segmenting the salient missing objects in a before eating image $I^b$ using information from a pair of before and after eating images, $I^b$ and $I^a$. We treat this problem as a paradigm of top down saliency detection where visual attention of HVS is guided by a task. Our proposed method uses $I^a$ as background to obtain a contrast map that is an estimate of the probability of pixels of $I^b$ belonging to objects/regions missing in $I^a$. The contrast map is then fused with the saliency information of $I^b$ to obtain a probability map  $\hat{M}^b_a$ for salient missing objects. 
Our experimental results validated that our approach achieves better performance both quantitatively and visually when compared to state-of-the-art salient object detection methods such as  R3NET\cite{r3net}, NLDF\cite{nldf}, Amulet\cite{amulet} and UCF\cite{ucf}. As discussed in Section~\ref{intro}, we have only considered the case where there is no food in the after eating image. In the future, we will extend our model to consider more general scenarios.

\bibliographystyle{ACM-Reference-Format}
\bibliography{ref}

\end{document}